\documentclass[preprint,12pt]{elsarticle}

\usepackage{amssymb}
\usepackage{colortbl}
\definecolor{mygray}{gray}{.9}
\definecolor{mygray2}{gray}{.8}

\usepackage{amsmath,amsfonts,amssymb}
\usepackage{algorithmic}
\usepackage{algorithmic}
\usepackage{bbding}
\usepackage{algorithm}
\usepackage{array}
\usepackage[caption=false,font=normalsize,labelfont=sf,textfont=sf]{subfig}
\usepackage{caption}
\usepackage{textcomp}
\usepackage{stfloats}
\usepackage{url}
\usepackage{verbatim}
\usepackage{graphicx}

\usepackage{multirow}
\usepackage{booktabs}

\journal{Nuclear Physics B}

\begin{document}

\begin{frontmatter}



\title{SARA: Controllable Makeup Transfer with Semantic-guided Alignment and Region-Adaptive normalization}


\author[inst1,inst2]{Xiaojing Zhong\fnref{fn}}

\affiliation[inst1]{organization={School of Software Engineering}, South China University of Technology}

\author[inst1]{Xinyi Huang\fnref{fn}}
\author[inst5]{Zhonghua Wu}
\author[inst2]{Guosheng Lin\corref{cor}}
\author[inst1,inst3,inst4]{Qingyao Wu\corref{cor}}

\fntext[co-first]{Equal Contribution}
\cortext[cor]{Corresponding authors}
\affiliation[inst2]{organization={School of Computer Science and Engineering}, Nanyang Technological University}
\affiliation[inst3]{organization={Key Laboratory of Big Data and Intelligent Robot},
country={Ministry of Education}}    
\affiliation[inst4]{Peng Cheng Laboratory}
\affiliation[inst5]{SenseTime Research}

\begin{abstract}
Makeup transfer is a process of transferring the makeup style from a reference image to the source image, while preserving the source image's identity. In addition to achieving fine-grained control over the makeup transfer, incorporating semantic alignment into the transformation is crucial, as the poses of the reference and source images are often inconsistent. We propose a novel Semantic-guided Alignment and Region-Adaptive normalization framework (SARA) to effectively transfer makeup styles under misaligned poses, offering flexible control to meet the demands of real-world applications, such as partial transfer, intensity adjustment, and makeup removal. Specifically, SARA comprises three modules: Firstly, we propose a semantic-guided alignment module to explicitly construct dense correspondence between the reference image and the target semantic map, employing unbalanced optimal transport matching to handle semantic region mismatches. Secondly, a region-adaptive normalization module is responsible for dynamically combining the warped style features with shape-independent style codes obtained by region-wise average pooling, mitigating feature loss during the alignment. Lastly, a makeup fusion module progressively fuses the identity features with the makeup styles to render the final output image. Furthermore, we combine optimal transport with histogram matching to generate the pseudo ground truth, which is used to facilitate the transfer in terms of both spatial alignment and color distribution. Experimental results show that our proposed SARA outperforms existing methods on two public datasets.

\end{abstract}

\begin{graphicalabstract}
\end{graphicalabstract}

\begin{highlights}
\begin{sloppypar}

\item We present SARA, a unified framework designed for controllable makeup transfer, which simultaneously supports partial transfer, degree-controlled transfer, and makeup removal.
\item To accurately transfer makeup styles under misaligned poses, we incorporate unbalanced optimal transport into semantic-guided feature alignment for explicitly constructing dense correspondence between mismatched semantic regions.
\item We propose to achieve region-adaptive normalization with shape-independent style codes to dynamically compensate for potential feature loss during the alignment process.
\end{sloppypar}
\end{highlights}


\end{frontmatter}


\section{Introduction}

Given a facial image with a specific makeup style, the makeup transfer task aims to apply that makeup style to a target face image while preserving the identity of the target face. This task has widespread applications in scenarios where users want to enhance their appearance virtually. Although it is easy to obtain makeup and non-makeup facial images from the internet, obtaining paired images of the same identity with different makeup styles is challenging, as it is rare to find absolutely identical faces with varying makeup styles.  To address this issue, one approach \cite{li2018beautygan,chang2018pairedcyclegan,choi2018stargan,huang2018multimodal} is to leverage the idea of CycleGAN \cite{zhu2017unpaired} constructing a cyclic training process involving two networks: one network transfers makeup styles from a source image to a target image, while the other network removes makeup from the transferred image. However, these methods treat different semantic regions of the face equally. This limitation hinders fine-grained control over the application of makeup transfer, which is an essential factor in achieving realistic and natural-looking transfer results.   

To achieve flexible and controllable makeup transfer, some methods \cite{jiang2020psgan,liu2021psgan++,sun2022ssat,yang2024makeup,xiang2022ramgan,deng2021spatially,sun2023ssat} encode the feature maps of the makeup image into learnable affine transformation parameters \cite{huang2017arbitrary}, which are then used to modulate the feature maps of the identity image. Due to the common issue of pose misalignment between the reference and source images, incorporating semantic alignment into the transformation is crucial. However, existing methods have limitations in effectively building semantic correspondence: \cite{jiang2020psgan,liu2021psgan++} require both facial landmarks and facial parsing masks to integrate spatial information into the attention matrix, leading to a relatively cumbersome process during model inference; \cite{deng2021spatially,yang2022elegant} fail to adequately capture the intricate details and spatial relationships within the makeup style due to their reliance on low-dimensional encoded vectors or sparse semantic correspondence; Although \cite{sun2022ssat} establishes dense semantic correspondences, its reliance on cosine similarity for aligning features results in many-to-one matching issues \cite{zhan2021unbalanced}, failing to accurately transfer in cases of semantic region mismatch.
\par 

\begin{figure}[t]
\centering
\setlength{\abovecaptionskip}{0pt}
\includegraphics[width=0.75\textwidth,height=0.6\textheight]{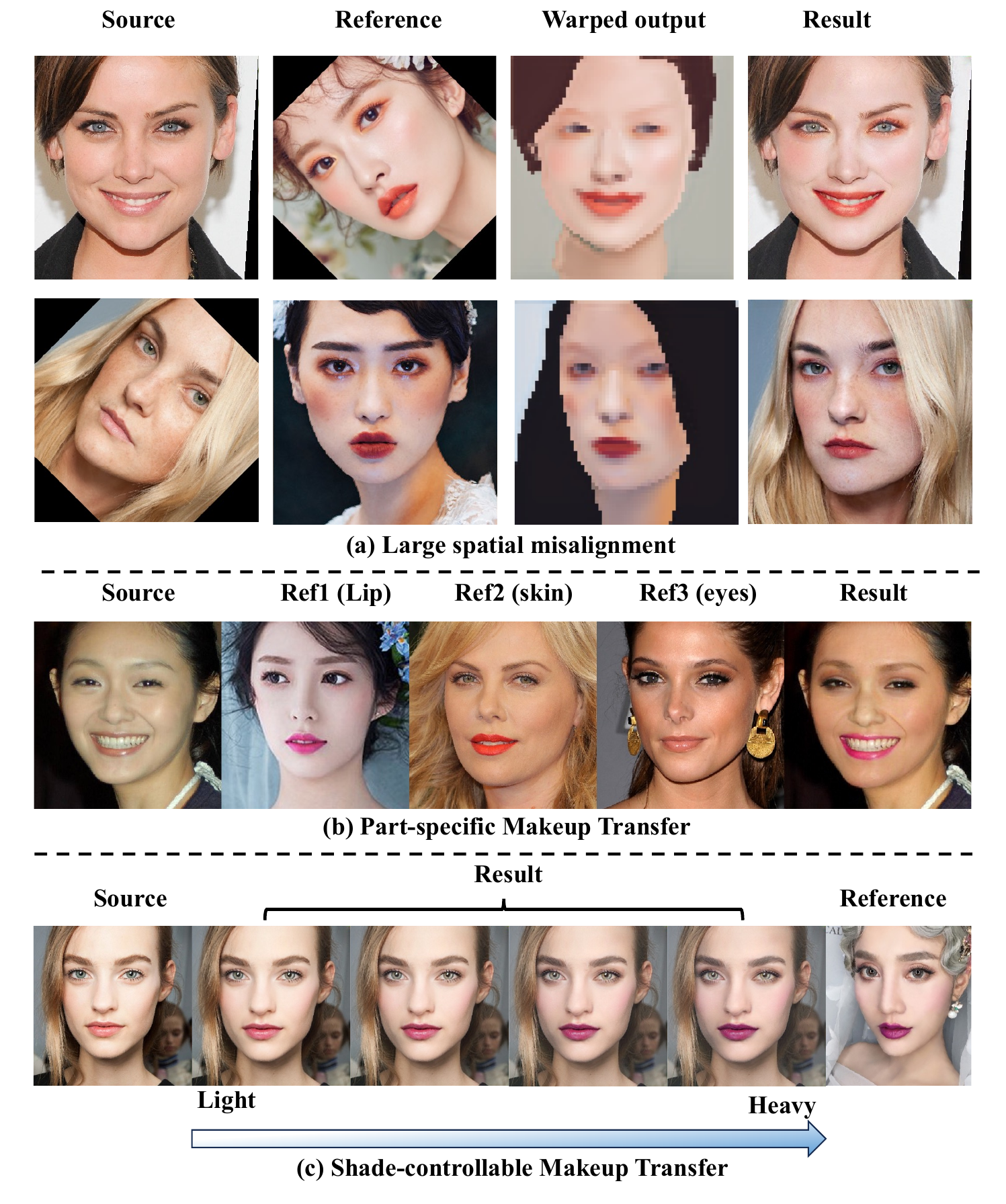}
\caption{\textbf{SARA supports flexible operations.} (a) SARA enables pose-robust transfer under the guidance of semantic alignment. (b) Users are allowed to select partial makeup styles from the reference image. (c) SARA can adjust the degree of the makeup styles. For best results, zoom in.}
\label{1}
\end{figure}

We propose a novel makeup transfer framework, Controllable Makeup Transfer with \textbf{S}emantic-guided \textbf{A}lignment and \textbf{R}egion-\textbf{A}daptive normalization (\textbf{SARA}), which has the capability to handle large spatial misalignment while enabling part-specific and degree-specific makeup transfer (see Fig. \ref{1}). First, we design a Semantic-guided Alignment Module (\textbf{SAM}) that warps the reference image based on the facial parsing map of the source image. We construct such cross-domain feature alignment to generate an intermediate result that simultaneously provides the reference styles and the source semantic shapes. Moreover, we employ unbalanced optimal transport for building dense correspondence during feature alignment, which effectively handles misaligned head poses, such as when the source identity requires makeup styles that are not present in the reference image. Given the coarseness of the intermediate results, we propose a Region-Adaptive normalization Module (\textbf{RAM}). This module extracts region-adaptive styles from the reference image and broadcasts them to the corresponding regions of the target semantic map, which compensates for potential feature loss during the alignment. Note that the target semantic map is obtained by warping using the optimal transport plan, rather than extracting the parsing map from the source image. The distinction lies in the fact that the transferred makeup styles coverage might not be consistent with the source identity's facial region. Finally, a Makeup Fusion Module (\textbf{MFM}) is responsible for allocating the modulated parameters to the fusion blocks. These parameters are dynamically combined from warped styles generated by SAM and shape-independent styles generated by RAM. MFM progressively fuses the identity features with the makeup styles to render fine-grained results.


\begin{table*}[t]
\tiny
    \centering
    \resizebox{1.0\linewidth}{!}{
    \begin{tabular}{c|c c c c c c}
    \toprule[0.5pt]
        \textbf{Property} &  BGAN & BGlow & LADN & EleGANt & SSAT &SARA(Ours) \\
    \bottomrule[0.pt]
    \toprule[0.5pt]
    Shade & \XSolidBrush & \CheckmarkBold & \CheckmarkBold & \CheckmarkBold & \CheckmarkBold &\CheckmarkBold  \\
     \bottomrule[0.pt]
   \toprule[0.5pt]
    Part & \XSolidBrush & \XSolidBrush &\XSolidBrush & \CheckmarkBold & \CheckmarkBold &\CheckmarkBold \\
     \bottomrule[0.pt]
       \toprule[0.5pt]
    Landmark-free & \CheckmarkBold & \CheckmarkBold & \XSolidBrush & \XSolidBrush & \CheckmarkBold &\CheckmarkBold \\
     \bottomrule[0.pt]
       \toprule[0.5pt]
    Misaligned & \XSolidBrush & \XSolidBrush & \XSolidBrush &\CheckmarkBold & \CheckmarkBold &\CheckmarkBold\\
     \bottomrule[0.pt]
       \toprule[0.5pt]

    Mismatch & \XSolidBrush & \XSolidBrush & \XSolidBrush & \CheckmarkBold & \XSolidBrush &\CheckmarkBold \\

   \bottomrule[0.5pt]
    \end{tabular}}
       \caption{$`Shade`$ refers to methods that allow for the control of makeup degree. $`Part`$ indicates the ability to perform partial transfer. $`Landmark-free`$ describes methods that do not require facial landmarks. $`Misaligned`$ refers to methods that can handle spatial misalignment between the source and reference images. $`Mismatch`$ indicates methods that can deal with semantic region mismatch between the source and reference images.}\label{tab:1p} 

\end{table*}


Tab. \ref{tab:1p} demonstrates the properties of various makeup transfer methods in summary. In brief, our main contributions are threefold:

\begin{itemize}
\begin{sloppypar}
\item{We present SARA, a unified framework designed for controllable makeup transfer, which simultaneously supports partial transfer, degree-controlled transfer, and makeup removal.}

\item{To accurately transfer makeup styles under misaligned poses, we incorporate unbalanced optimal transport into semantic-guided feature alignment for explicitly constructing dense correspondence between mismatched semantic regions.}

\item{We propose to achieve region-adaptive normalization with shape-independent style codes to dynamically compensate for potential feature loss during the alignment.}

\end{sloppypar}
\end{itemize}

\section{Related Work}

\subsection{Makeup Transfer}

Makeup transfer has garnered significant attention over the last decade \cite{wang2016face,li2015simulating,li2018anti,guo2009digital}. CycleGAN \cite{zhu2017unpaired} can be applied to facial images to transfer general makeup styles by learning domain-to-domain translation from two sets of images, one with makeup and the other without. PairedCycleGAN \cite{chang2018pairedcyclegan} builds upon CycleGAN by introducing a paired cycle GAN specifically designed for makeup transfer and removal, enabling the transfer of makeup from a reference image to a target face. Li $\emph{et~al.}$ propose to constrain the local facial regions by matching the color histograms of corresponding areas between the reference and target images \cite{li2018beautygan}. \cite{nguyen2021lipstick,gu2019ladn} focus on transferring dramatic makeup styles. Additionally, Lyu $\emph{et~al.}$ introduce a 3D-aware GAN that unwarps the facial texture and refines it based on facial symmetry to handle shadows and occlusions \cite{lyu2021sogan}. Wan $\emph{et~al.}$ improve the fidelity of local regions sensitive to color by consolidating the transformer architecture into makeup transfer \cite{wan2022facial}
\par
However, these methods lack user control over the transfer process. To address this, Chen $\emph{et~al.}$ propose BeautyGlow \cite{chen2019beautyglow}, which disentangles makeup style and facial identity of latent vectors, enabling users to adjust the intensity of the transferred style. PSGAN \cite{jiang2020psgan} and improved PSGAN++ \cite{liu2021psgan++} propose to handle the spatial misalignment problem facing non-frontal images by incorporating visual appearances and locations into the attention mechanism. EleGANt \cite{yang2022elegant} uses a pyramid structure with a high-resolution feature map to preserve high-frequency makeup features beyond color distributions, but it lacks careful consideration of large misalignment. Nevertheless, both methods lack careful consideration of large misalignment and rely heavily on facial landmarks, which introduces additional computational overhead. SCGAN \cite{deng2021spatially} separates facial component into low-dimensional feature vectors according to semantic layouts, which are recombined in a specific order to render the image. Although \cite{sun2022ssat} establishes a dense correspondence to perform semantic alignment, it does not effectively handle the problem of semantic region mismatch that arises when the poses between the source and reference images are misaligned.

\subsection{Style Transfer}

Conventional style transfer methods rely on hand-crafted algorithms to render an image in fixed styles \cite{haeberli1990paint,hertzmann1998painterly}, or match two images with hand-crafting features \cite{hertzmann2001image,efros2001image}. More recent advances in style transfer have shifted towards the use of deep learning techniques, specifically the increasingly effective generative models \cite{tang2023datfuse,zhang2024real,zhou2022hrinversion,zhou2023cips,yin20233d}. AdaIN \cite{huang2017arbitrary} and DIN \cite{jing2020dynamic} use conditional instance normalization to align content and style feature statistics, but dynamic generation of affine parameters may cause distortion artifacts. Other methods use an autoencoder-based framework with feature transformation and/or fusion \cite{dumoulin2016learned,zhang2023inversion,zhang2023inversion,deng2022stytr2}. These models have shown impressive results in generating realistic images in a given style or transferring the style of one image onto another \cite{chen2022quality}. Controlling the output of style transfer is crucial for downstream applications. One approach to achieve this is through spatial style control, which allows users to selectively apply features from specific regions of the style image to designated areas of the output \cite{lu2017decoder,kolkin2019style}. Another approach involves global style control methods that decompose the overall attributes of images, such as hue, saturation, and illumination \cite{gatys2017controlling}. Compared to general style transfer tasks, makeup transfer requires a higher level of precision and fine-grained control over specific facial regions, which leads to a discrepancy between the generated and desired results when applying these techniques to makeup transfer.



\begin{figure}[t]
\centering
\setlength{\abovecaptionskip}{0pt}
\includegraphics[width=1.0\textwidth,height=0.43\textheight]{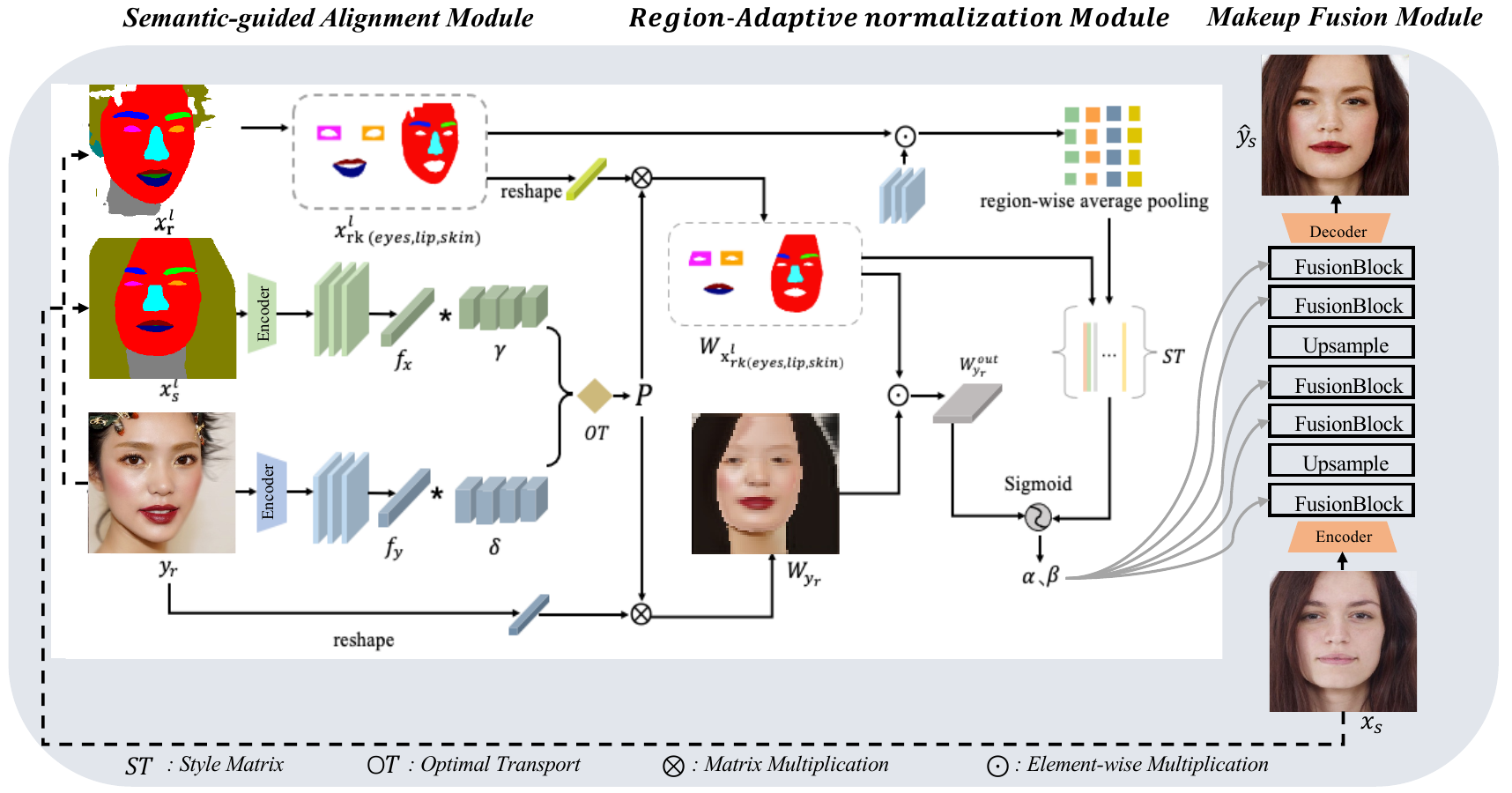}
\caption{\textbf{Overview of our proposed method.} It mainly has three modules: (i) Semantic-guided alignment module warps the reference image $y_r$ and the partial reference semantic map $x_{rk}^l$ to be aligned with the source semantic map $x_s^l$ through estimating the dense correspondence between $y_r$ and $x_s^l$, where $i=\{lip,skin,eyes\}$. (ii) Region-adaptive normalization module decouples the shape and style of makeup via a region-wise average pooling layer, broadcasting the shape-independent style codes to the target semantic map $W_{x_{rk}^l}$ to generate the style matrix $ST$. The modulated parameters are dynamically combined with $W_{y_r}^{out}$ and $ST$.(iii) Makeup fusion module progressively fuses makeup styles with the identity features, to generate the fine-grained result $\hat{y}_s$.}
\label{2}
\end{figure}

\section{Methodology}

\subsection{Problem Formulation and Notations}

Let X and Y denote the domain of the images with non-makeup and those with makeup, respectively. Given a source image $x_s \in X$ and a reference image $y_r \in Y $, our goal is to learn a mapping function: $f(x_s,y_r \rightarrow \hat{y}_s)$, where $\hat{y}_s$ has the same makeup style as $y_r$ while preserving the facial identity of $x_s$. Furthermore, as makeup removal is a particular case of makeup transfer, we also learn a mapping function: $\hat{f}(y_r,x_s \rightarrow \hat{x}_r)$, where $\hat{x}_r$ has the same makeup style as $x_s$ while preserving the facial identity of $y_r$.  

\subsection{Network Structure}

The overall framework of SARA is shown in Fig. \ref{2}, which consists of three modules: a Semantic-guided Alignment Module \textbf{(SAM)}, a Region-Adaptive normalization Module \textbf{(RAM)} and a Makeup Fusion Module \textbf{(MFM)}. The content of each module will be introduced in detail (Sec \ref{3.2.1} $\sim $ Sec \ref{3.2.3}). The objective functions of the whole model will be described in Sec \ref{3.3}.


\subsubsection{Semantic-guided Alignment Module}\label{3.2.1} 

To preserve detailed makeup styles with spatial context, we warp the reference image that provides the desired makeup styles by explicitly building dense correspondence to eliminate the effect induced by pose misalignment. Inspired by \cite{zhan2021unbalanced}, we propose to address the feature alignment problem using unbalanced optimal transport. Optimal transport aims to determine a transport plan that transports samples from one distribution to another with minimal cost. However, when the source and reference images have different poses, such as frontal and side images, the total masses of the two distributions are often unequal. We leverage unbalanced optimal transport with divergence metric constraints to handle this problem. 
\par
As shown in Fig. \ref{2}, SAM starts with a reference image $y_r$ and a source semantic map $x_s^l$, which is obtained by running an off-the-shelf face parsing network \cite{wei2017learning} on $x_s$. We employ two independent feature extraction operators to extract feature vectors from $x_s^l$ and $y_r$, yielding corresponding sets $f_x=\{s_1,...,s_n\}$ and $f_y=\{r_1,...,r_n\}$, respectively, with $n$ denoting the number of the feature vectors. We denote the mass corresponding to $s_i$ and $r_j$ as $\gamma_i$ and $\delta_j$, respectively, where $i,j \in [1,n]$. The total masses of two sets are $\gamma = \sum_{i=1}^n\left(\gamma_i\right)$ and $\delta = \sum_{j=1}^n\left(\delta_j\right)$. Given that the mass assigned to a feature vector should depend on its similarity with the other feature set, we dynamically determine $\gamma_i$ and $\delta_j$ by:

\begin{equation}
    \gamma_i=s_i \cdot \frac{\sum_{i=1}^n\left(r_i\right)}{n}, \delta_j=r_j \cdot \frac{\sum_{j=1}^n\left(s_j\right)}{n}.
\end{equation}

Each entry $C_{ij}$ in the distance matrix $C$ represents the cost of transporting mass $\gamma_i$ to mass $\delta_j$, which can be computed by $C_{i j}=1-\frac{s_i^{\top} \cdot r_j}{\left\|s_i\right\|\left\|r_j\right\|}$. To address the unbalanced transport problem, we incorporate Kullback-Leibler (KL) divergence into the optimal transport, where $\mathrm{KL}(a \| b)$ is given by $\sum_{i=1}^n a_i \log \left(\frac{a_i}{b_i}\right)-a_i+b_i$. In addition, the optimal transport problem is not differentiable everywhere due to its linear objective and constraints. To make it strictly convex and differentiable, we add an entropic regularization term $H(P)=-\sum_{i, j=1}^n P_{i j} \log P_{i j}$, where $P$ represents the transport matrix with $P_{i,j}$ denoting the amount of masses transported between $\gamma_i$ and $\delta_j$. The optimal transport can be formulated as follows: 

\begin{equation}\label{eq1}
\min _P\left[\sum_{i, j=1}^n C_{i j} P_{i j}+\tau \mathrm{KL}(P \overrightarrow{1} \| \gamma)+\tau \mathrm{KL}\left(P^{\top} \overrightarrow{1} \| \delta\right)-\eta H(P)\right],
\end{equation}

where $\tau$ is the regularization parameter for KL divergence terms and $\eta$ is the regularization coefficient that regulates the smoothness and dispersion of the transport plan.

To solve the entropic unbalanced optimal transport problem efficiently, we represent Eq. \ref{eq1} in the Fenchel-Legendre dual form:

\begin{equation}\label{eq3}
\begin{aligned}
 \max _{u, v}\left[-M^*(-u)-N^*(-v)-\eta \sum_{i, j} \exp \left(\frac{u_i+v_j-C_{i j}}{\eta}\right)\right], where\\
M^*(u)=\max _s s^{\top} u-\tau \mathrm{KL}(s \| \delta), N^*(v)=\max _r r^{\top} v-\tau \mathrm{KL}(r \| \gamma).
\end{aligned}
\end{equation}

We employ Sinkhorm algorithm \cite {cuturi2013sinkhorn} to tackle the solution of Eq. \ref{eq3}, which enables us to obtain an optimal transport plan $P$. $P$ is encoded by the dual vectors $u$ and $v$ and can be expressed as follows:
\begin{equation}
P_{i j}=\gamma_i \delta_j \exp \frac{1}{\eta}\left[u_i+v_j-C_{i j}\right].
\end{equation}

\begin{figure}[t]
\centering
\setlength{\abovecaptionskip}{0pt}
\includegraphics[width=0.75\textwidth,height=0.3\textheight]{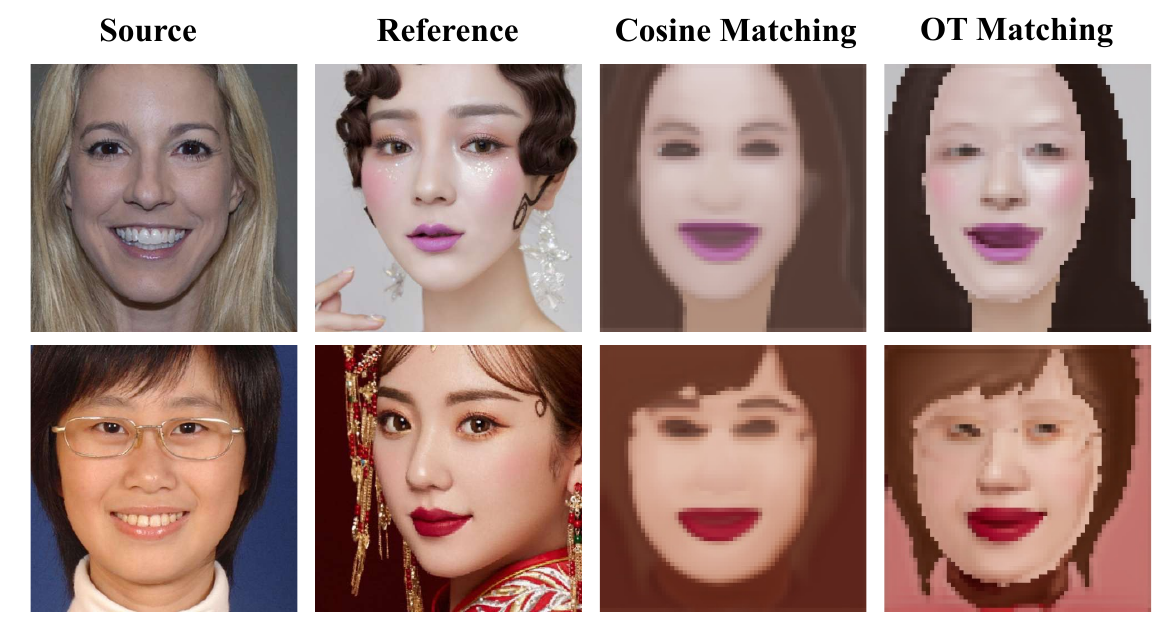}
\caption{\textbf{The comparison of warped results with different matching.} The third column depicts cosine matching results, while the fourth shows those from OT matching. The warped results generated from OT matching preserve more intricate makeup features, such as blusher.}
\label{3}
\end{figure}

To perform controllable makeup transfer, we extract specific components of facial regions that generally embody makeup styles. Specifically, given the optimal transport plan $P$, we warp the reference image $y_r$ to obtain the intermediate result $W_{y_r}$ that preserves the source identity's shape but exhibits the desired makeup styles. We also warp the reference semantic map $x_r^l$ to extract specific components of $W_{y_r}$, denoted as $W_{x_{rk}^l}$, where $k=\left\{lip,skin,eyes\right\}$. This can be represented as: 
\begin{equation}
\begin{aligned}
W_{y_r} & = y_r \cdot P,\\
W_{x_{rk}^l} & = x_r^l \cdot P,
\end{aligned}
\end{equation}
The filtered makeup regions can be obtained as follows:
\begin{equation}\label{eq6}
W_{y_r}^{out}= W_{x_{rk}^l} \odot W_{y_r},
\end{equation}
where $\odot$ denotes element-wise multiplication for each pixel. Besides, $W_{x_{rk}^l}$ is also utilized to aid in the target semantic map generation in the following per-region normalization. Since SAM models global correspondence, it can effectively warp makeup regions that cover large areas, such as blush. Fig. \ref{3} showcases that cosine similarity matching produces smooth results, while the warped results from optimal transport matching contain more fine-grained makeup styles.

\subsubsection{Region-Adaptive normalization Module}\label{3.2.2}

As styles can be considered shape-independent embedded codes that modulate the affine transformation parameters of normalization layers \cite{deng2021spatially,zhu2020sean,zhang2021pise,lv2021learning}, we dynamically combine the region-adaptive style codes with the coarse warped output from SAM,
which enhances the sharpness and detail of the transferred styles. Inspired by \cite{zhu2020sean}, we takes the region-specific style codes extracted from the reference features and then broadcast them to the target semantic regions on facial parts. Rather than re-extracting features from the reference image $y_r$, we adopt the reference features obtained from the alignment module as input. Intuitively, we contend that this could serve in maintaining the consistency and continuity of the original features since using external data may break it. 
\par

\begin{figure}[t]
\centering
\setlength{\abovecaptionskip}{0pt}
\includegraphics[width=0.7\textwidth,height=0.3\textheight]{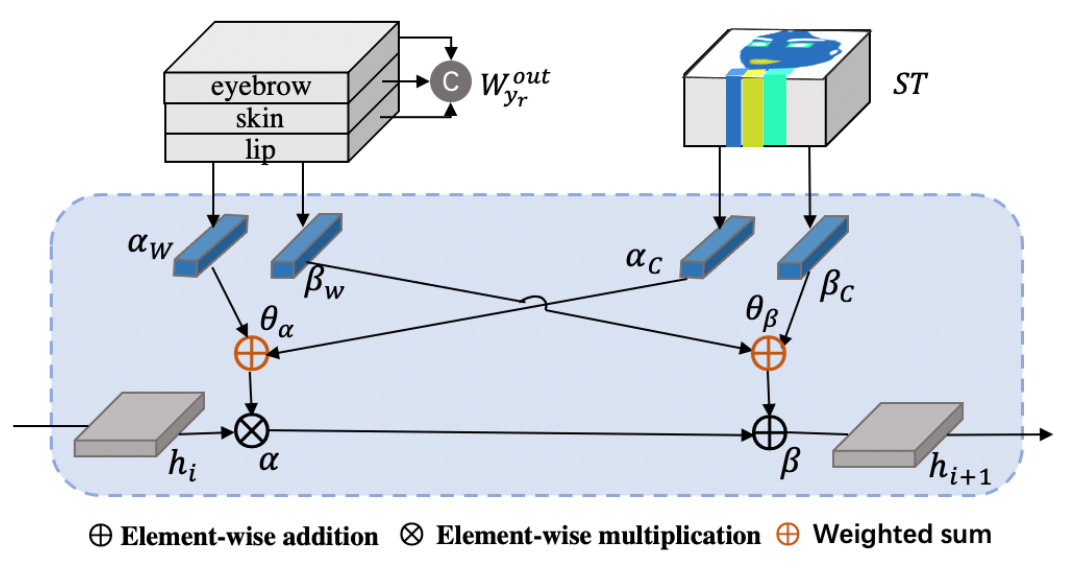}
\caption{\textbf{Dynamic combination of region-adaptive normalization.} The scale and bias parameters $\alpha$ and $\beta$ are weighted from the warped out $W^{out}_{y_r}$ and the style matrix $ST$.}
\label{4}
\end{figure}

We aggregate shape-independent style codes through region-wise average pooling into a style matrix $ST\in \mathbb{R}^{256 \times s}$, where $s$, being 3 in our case, represents the dimensional concatenation of lip, skin and eye shadow. With the warped semantic map $W_{x_{rk}^l}$ as guidance, we broadcast the style codes to their corresponding target regions, generating the target style map. Following \cite{zhu2020sean}, we can achieve controllable normalization using solely the style map. In our paper, we propose to dynamically combine it with the warped output $W_{y_r}$ generated from SAM, which preserves the spatial context of makeup styles. We send $W_{y_r}$ and $ST$ to separate convolutional layers to learn two sets of modulation parameters. To be specific, let $h^i \in \mathbb{R}^{B \times C^i \times H^i \times W^i} $ denote the activation values before the i-th normalization layer, where $B$,$C^i$,$H^i$,$W^i$ represent the batch size, number of channels, height and width of the i-th feature map, respectively. The mean $\mu^i_{c}$ and standard deviation $\sigma^i_{c}$ are calculated in the channel-wise manner. Then, the output of RAM in the i-th layer is given by:
\begin{equation}
h^{i+1}_{b,c,h,w}= \alpha^i_{c,h,w}\frac{h_{b,c,h,w}^i-\mu^i_{c}}{\sigma^i_{c}}+\beta^i_{c,h,w},
\end{equation} 
where $\alpha^i_{c,h,w}$ and $\beta^i_{c,h,w}$ are the learnable parameters. As shown in Fig. \ref{4}, they are the weighted sum of these features from the convolution output of the combined warped out $W_{y_r}^{out}$ and the style matrix $ST$, which are defined as:
\begin{equation}
\begin{aligned}
\alpha^i_{c,h,w} = \theta_{\alpha}\cdot \alpha_W+(1-\theta_{\alpha})\cdot \alpha_C \\
\beta^i_{c,h,w} = \theta_{\beta}\cdot \beta_W+(1-\theta_{\beta})\cdot \beta_C,
\end{aligned}
\end{equation} 
where $\theta_{\alpha}$ and $\theta_{\beta}$ are the adaptation coefficients that are dynamically weighted from two sources in the transfer. Different from \cite{zhu2020sean}, which weights the parameters $\theta_{\alpha}$ and $\theta_{\beta}$ using the style map and the semantic mask, our approach incorporates spatial alignment into the normalization to adaptively modulate the activation values. 

\subsubsection{Makeup Fusion Module}\label{3.2.3}

MFM consists of a face identity encoder, two upsampling layers, five fusion blocks, and a decoder. Unlike \cite{zhu2020sean,park2019arbitrary}, which begins with random noises, our approach aims to preserve the source identity while transferring the makeup style. We utilize a face identity encoder to extract the identity features from the source image $x_s$. The extracted features are then processed by a series of fusion blocks, which receive scale and bias parameters from RAM to denormalize the feature maps. We employ upsampling layers that gradually increase the spatial dimensions of the feature maps. Finally, the decoder takes the output of the last fusion block as input to generate the transferred image $\hat{y}_s$, which combines the makeup style from the reference image with the identity features of the source image.

\subsection{Loss Functions}\label{3.3}

The losses used to constrain the optimal transport plan of SAM include a domain alignment loss \cite{zhang2020cross}, which ensures that the features extracted by two independent feature extraction operators lie in the same domain, and a cycle-consistency loss which warps the warped output back to the original domain using the optimal transport plan, with the expectation that this twice-warped image should be identical to the reference image. Next, we denote the loss for jointly training RAM and MFM as $\mathcal{L}_{joint}$. $\mathcal{L}_{joint}$ can be defined as follows:

\begin{equation}
\begin{aligned}
\mathcal{L}_{joint}= \lambda_{1}\mathcal{L}_{perc} +\lambda_{2}\mathcal{L}_{makeup}+\lambda_{3}\mathcal{L}_{cycle}+\lambda_{4}\mathcal{L}_{adv}+\lambda_{5}\mathcal{L}_{id},
\end{aligned}
\end{equation} 
where $\lambda_{1,2,3,4,5}$ are the hyper-parameters to control the weights of each term.

\textbf{Perceptual loss.} We utilize the VGG-19 pretrained model \cite{simonyan2014very} to extract features from the generated output $\hat{y}_s$ and the input source image $x_s$, respectively. Since $\hat{y}_s$ and $x_s$ are not aligned at the pixel-level, we only use the activation after the $relu4\_2$ layer, which is denoted as $\phi_l$. $\mathcal{L}_{perc}$ can be formulated as below: 
\begin{equation}
\mathcal{L}_{perc} = ||\phi_l(\hat{y}_s)-\phi_l(x_s)   ||_2,
\end{equation}
where $||\cdot||_2$ is the L2-Norm.

\begin{figure}[t]
\centering
\setlength{\abovecaptionskip}{0pt}
\includegraphics[width=0.75\textwidth,height=0.3\textheight]{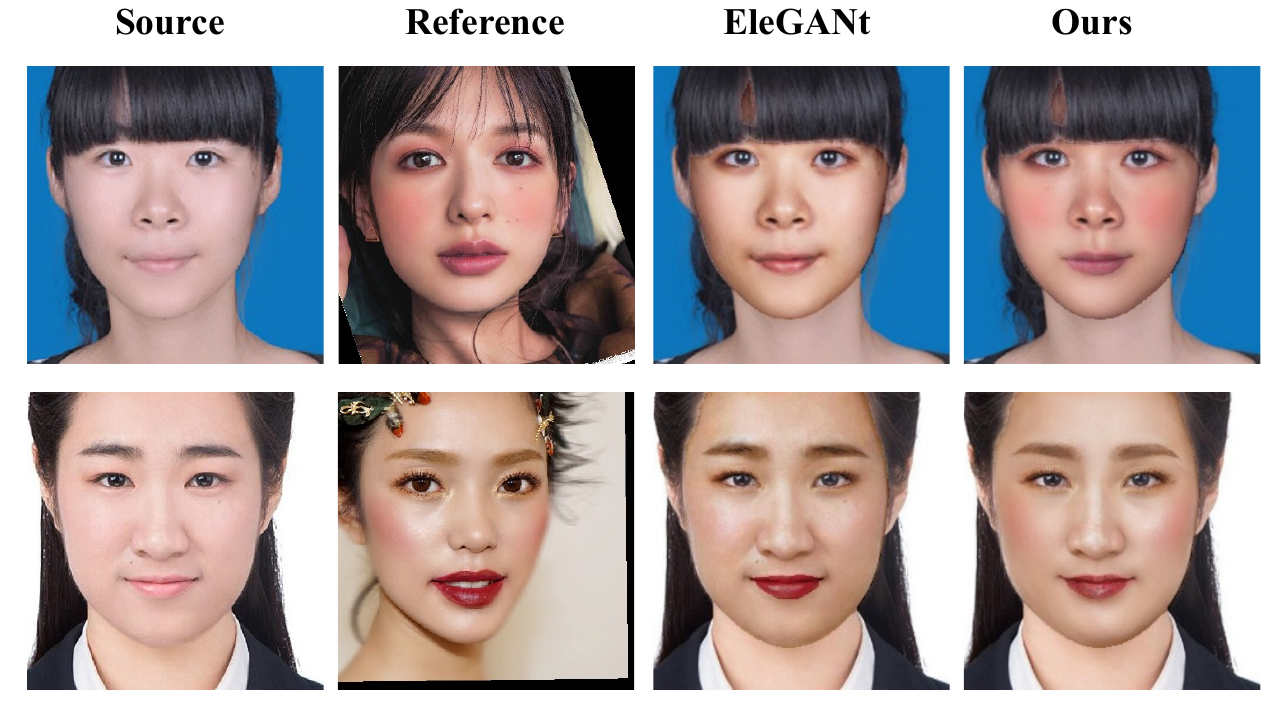}
\caption{\textbf{Comparison of generated pseudo ground truth between EleGANt \cite{yang2022elegant} and ours.}}
\label{5}
\end{figure}

\textbf{Makeup loss.} \cite{yang2022elegant} proposes generating pseudo ground truth based on Thin Plate Spline (TPS) warping \cite{wan2022facial} and histogram matching. However, TPS warping with a low degree of freedom is insufficient to preserve intricate makeup styles under large geometric changes. To address this limitation, we improve the generation of pseudo ground truth by replacing TPS warping with optimal transport matching as it establishes dense correspondences between distributions. Specifically, we employ histogram matching to align the color distribution of the lip, skin and eye shadow regions in the source image with those of the reference image. This ensures that the transferred makeup style closely resembles the desired appearance in terms of color and intensity. Then, we extract the corresponding regions from the warped result generated by optimal transport matching and blend them with the output of the histogram matching to generate the final pseudo ground truth. As illustrated in Fig. \ref{5}, the pseudo ground truth generated by our method preserves accurate large-area blush. Let $PGT(x,y)$ and $PGT(y,x)$ denote the generated pseudo ground truth for makeup transfer and inverse transfer, respectively. The makeup loss $\mathcal{L}_{makeup}$ can be formulated as below: 
\begin{equation}
\begin{aligned}
\mathcal{L}_{makeup} & = ||f(x_s,y_r)-PGT(x_s,y_r)||_2 \\
&  + ||f(y_r,x_s)-PGT(y_r,x_s)||_2
\end{aligned}
\end{equation}


\textbf{Cycle consistency loss.} We employ a global cycle loss \cite{zhu2017unpaired} to enforce the consistency of the mapping between two domains. $\mathcal{L}_{cycle}$ can be calculated as below: 
\begin{equation}
\begin{aligned}
\mathcal{L}_{cycle} & = ||f(f(y_r,x_s),y_r)-y_r||_1 \\
&  + ||f(f(x_s,y_r),x_s)-x_s||_1
\end{aligned}
\end{equation}

\textbf{Identity loss.} We construct the identity loss by referring to the reference, source, and output images as the same images. This aims to maintain consistency of the identity during the generation. $\mathcal{L}_{id}$ can be formulated as below:

\begin{equation}
\begin{aligned}
\mathcal{L}_{id} & = ||f(x_s,x_s)-x_s||_1 \\
&  + ||f(y_r,y_r)-y_r||_1
\end{aligned}
\end{equation}



\par
\textbf{Adversarial loss.} We adopt a discriminator \cite{mirza2014conditional} to constrain the latent space of output into the data distribution like $y_r$ in order to improve the quality of the generated image $\hat{y}_s$. $\mathcal{L}_{GAN}^G$ and $\mathcal{L}_{GAN}^D$ are respectively formulated as below: 
\begin{equation}
\begin{aligned}
\mathcal{L}_{GAN}^D & =-\mathbb{E}[h(D(y_r))]-\mathbb{E}[h(-D(G(x_s,y_r)))] \\
\mathcal{L}_{GAN}^G &  = -\mathbb{E}[D(G(x_s,y_r))],
\end{aligned}
\end{equation}
where the hinge function $h(t)=min(0,-1+t)$ is expected to regularize the discriminator \cite{zhang2019self,brock2018large}.
\par

\section{Experiments}



\subsection{Implementation Setting and datasets}

\textbf{Datasets.} We train our model using the Makeup Transfer (MT) dataset \cite{li2018beautygan}, which contains 2719 makeup images and 1115 non-makeup images, each depicting a different human subject with variations in poses, facial expressions. Following \cite{li2018beautygan}, we randomly select 250 makeup images and 100 non-makeup images from the MT dataset to form the testing set, while the remaining images are used for training. To further validate the effectiveness of our model on facial images with complex background, we also evaluate its performance on the M-wild dataset \cite{jiang2020psgan}. 

\par 

\textbf{Implementation Details.} To address the lack of direct supervision for optimal transport, we formulate the problem as a mask-to-image translation task by treating all non-makeup images, makeup images, and their corresponding semantic maps as inputs and outputs. We train SAM by integrating it with a translation network \cite{zhan2021unbalanced}. Once the training is complete, we discard the translation network and keep the semantic-guided warping module fixed. Using the intermediate results generated by SAM, we jointly train RAM and MFM to render the final image. All the experiments were conducted on one NVIDIA GeForce RTX 3090 GPU, and it takes roughly 2 days to train the whole model when batch size is set to 1. We empirically set the hyper-parameters of formula 9 as $\big(\lambda_1$=0.001,$\lambda_2$=\{0.1 (skin),1.5 (eyes),1.0 (lip)\},$\lambda_3$=1.0,$\lambda_4$=10.0,$\lambda_5$=1.0)$\big)$.


\begin{figure*}
\centering
\setlength{\abovecaptionskip}{0pt}
\includegraphics[width=1.0\textwidth,height=0.35\textheight]{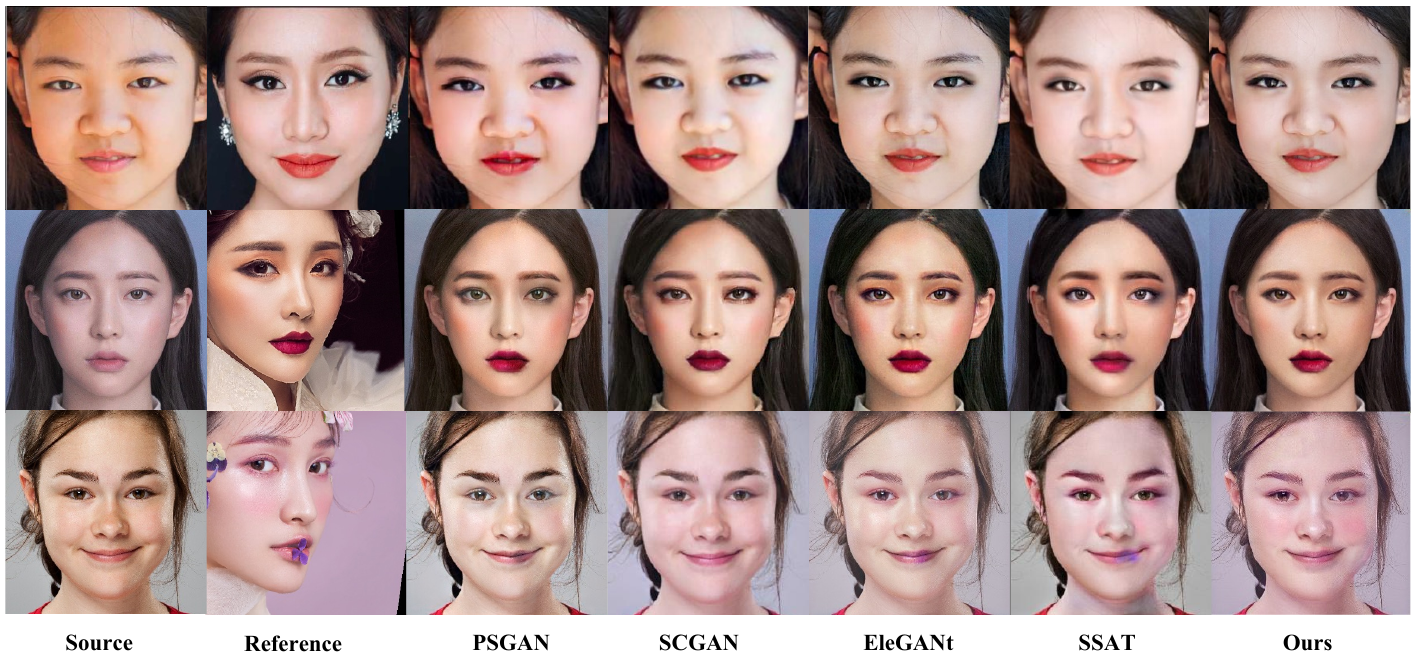}
\caption{\textbf{Qualitative comparison in MT dataset.} The compared methods include: \textbf{PSGAN} [CVPR'{\color{red}20}], \textbf{SCGAN} [CVPR'{\color{red}21}], 
\textbf{EleGANt} [ECCV'{\color{red}22}],
\textbf{SSAT} [AAAI'{\color{red}22}].} 
\label{6}
\end{figure*}

\subsection{Qualitative Results}\label{4.2}

\begin{figure*}
\centering
\setlength{\abovecaptionskip}{0pt}
\includegraphics[width=1.0\textwidth,height=0.25\textheight]{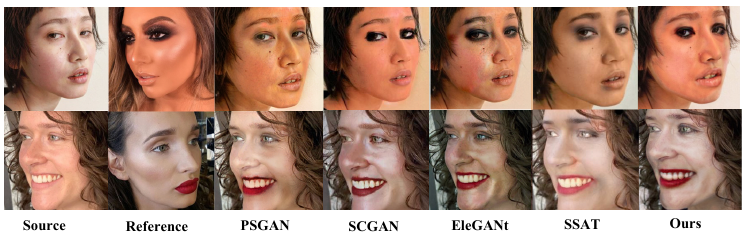}
\caption{\textbf{Qualitative comparison on test samples with large spatial misalignment in M-Wild dataset.} The compared methods include:  \textbf{PSGAN} [CVPR'{\color{red}20}], \textbf{SCGAN} [CVPR'{\color{red}21}], 
\textbf{SSAT} [AAAI'{\color{red}22}],
\textbf{EleGANt} [ECCV'{\color{red}22}].} 
\label{7}
\end{figure*}

We qualitatively compare our method with PSGAN \cite{jiang2020psgan}, SCGAN \cite{deng2021spatially}, EleGANt \cite{yang2022elegant} and SSAT \cite{sun2022ssat}. Fig. \ref{6} presents the comparative results tested on the MT dataset, where the pose difference between the makeup and non-makeup images is relatively small, and the makeup styles are generally light. PSGAN generates inaccurate transfer results particularly for the eye shadow region when the reference image is not a frontal face. Moreover, it fails to automatically complete occluded areas during the transfer, such as the partially obscured lips in the third row of Fig. \ref{6}. SCGAN alters the background color and fails to precisely transfer the shape of the eye shadow. Although EleGANt can transfer eye shadow, it fails to produce the desired blush effect in the transferred result when the source image lacks blush, even if the reference image exhibits blush. SSAT yields unsatisfactory results when dealing with misaligned poses and occlusions, failing to accurately transfer both the eye shadow and the lipstick. Our method effectively transfers the eye shadow, lipstick, and blush, handling cases with pose misalignments and occlusions between the source and reference images. To further validate the performance of each method under large pose misalignments, we also test them on the M-Wild dataset. As shown in Fig. \ref{7}, when facing these complex samples, the most noticeable issue is the generation of uneven skin tones, as observed in the results of SCGAN and EleGANt. Moreover, when dealing with dramatic makeup styles, methods like SSAT tend to produce artifacts. On the other hand, PSGAN barely achieves any transfer of dramatic eye shadow. In contrast, our method effectively handles shadows during the transfer and adaptively transfers dramatic eye shadow to the target face.

\begin{table*}[t]
\tiny
    \centering
    \resizebox{1.0\linewidth}{!}{
    \begin{tabular}{c c|c c c c c}
    \toprule[0.5pt]
  \multicolumn{2}{c|}{\textbf{User Study $\uparrow$}}
   &  PSGAN & SCGAN & EleGANt & SSAT  & SARA(Ours) \\
    \bottomrule[0.pt]
    \toprule[0.5pt]
    \multirow{2}{*}{Aligned} & Transfer-Accuracy & 2.95 & 2.61 & 2.99 & 3.08 & \textbf{3.37} \\
	& Image-Quality  & 2.93 & 2.96 & 3.02  & 2.97 & \textbf{3.12} \\ 
  \bottomrule[0.pt]
    \toprule[0.5pt]
     \multirow{2}{*}{Misaligned } & Transfer-Accuracy & 2.73 & 2.97 & 3.23 & 2.62 & \textbf{3.45} \\
	& Image-Quality  & 2.85  & 2.88 & 3.07  & 2.98 & \textbf{3.22} \\ 
    
   \bottomrule[0.5pt]
    \end{tabular}}
    \caption{Quantitative evaluation of the results obtained from different approaches, including PSGAN \cite{jiang2020psgan}, SCGAN \cite{deng2021spatially}, EleGANt \cite{yang2022elegant}, SSAT \cite{sun2022ssat}, and SARA. $`Aligned`$ and $`Misaligned`$ correspond to whether the poses of the source image and the reference image are aligned or misaligned, respectively. The best scores are highlighted in \textbf{bold}.}\label{compare} 
\end{table*}

\subsection{Quantitative Results}

As there are no well-established quantitative metrics for evaluating the quality of makeup transfer results, we follow previous methods \cite{sun2022ssat,yang2022elegant,deng2021spatially} to conduct user studies to assess the overall quality of the generated images. In our case, we evaluate the model's performance in terms of transfer accuracy and the quality of generated images on two types of test samples: aligned and misaligned, based on the pose discrepancy between the source and reference images. For each type of sample, we present 15 sets of images to 20 volunteers. Each set includes a source image, a reference image, the corresponding text prompt, and five randomly ordered transfer results generated by PSGAN \cite{jiang2020psgan}, SCGAN \cite{deng2021spatially}, EleGANt \cite{yang2022elegant}, SSAT \cite{sun2022ssat}, and our proposed method, SARA. The volunteers are asked to rank the five transfer results based on transfer accuracy and image quality, using a scoring system ranging from 5 (highest) to 1 (lowest), with no repeated scores allowed within each set. The final report will present the average scores across all sets. From Tab. \ref{compare}, our approach attains the top human preference scores, whether in terms of transfer accuracy or image quality.

\par 



\subsection{Controllable Makeup Transfer}

\subsubsection{Partial Makeup Transfer}

The makeup styles used for transfer originate from two sources: the aligned warped output $W_{y_r}^{out}$ and the non-aligned style matrix $ST$, enabling extraction of specific regions from different reference images for partial makeup transfer. We can reformulate Eq. \ref{eq6} as below:

\begin{equation}
\begin{aligned}
W_{y_r}^{out} & =W_{x_r^l\left\{eyes\right\}}\odot W_{y_r\left\{eyes\right\}}+ W_{x_r^l\left\{skin\right\}}\odot W_{y_r\left\{skin\right\}} \\
& + W_{x_r^l\left\{lip\right\}}\odot W_{y_r\left\{lip\right\}},
\end{aligned}
\end{equation} 

where $\odot$ denotes element-wise multiplication. For the three components in the above equation, $W_{y_r}^{out}$ and $W_{x_{r}^l}$ can be sourced from different reference images. Since $ST$ is shape-independent, we can concatenate $ST$ derived from region-wise average pooled feature maps across different reference images, then broadcast the concatenated $ST$ to the combined warped semantic map. Fig. \ref{1}(b) showcases the partial makeup transfer result when the lip styles originate from \textit{Ref 1}, the skin styles from \textit{Ref 2}, and the eye styles from \textit{Ref 3}.

\subsubsection{Shade-controllable Makeup Transfer}

We can perform linear interpolation on $W_{y_r}^{out}$ and $ST$ to control the degree of makeup style transfer. This operation can be formulated as:
\begin{equation}
\begin{aligned}
W_{y_r}^{out} =\epsilon W_{y_{r1}}^{out}+(1- \epsilon) W_{y_{r2}}^{out} \\
ST=\epsilon ST_1+(1- \epsilon) ST_2
\end{aligned}
\end{equation} 

The coefficient $\epsilon\in[0,1]$ in the above formula controls the degree of makeup styles. Specifically, when $W_{y_{r1}}^{out}$ comes from the reference image and $W_{y_{r2}}^{out}$ comes from the source image, the control of the makeup style degree is unidirectional. However, when both $W_{y_{r1}}^{out}$ and $W_{y_{r2}}^{out}$, and even more warped results, are derived from different reference images, interpolation can be performed on multiple makeup styles. As Fig. \ref{1}(c) shows, when we vary $\epsilon$ from 0 to 1 with an increment of 0.2, we can obtain global transfer results with varying degrees of intensity. 
\par

\begin{figure*}
\centering
\setlength{\abovecaptionskip}{0pt}
\includegraphics[width=1.0\textwidth,height=0.35\textheight]{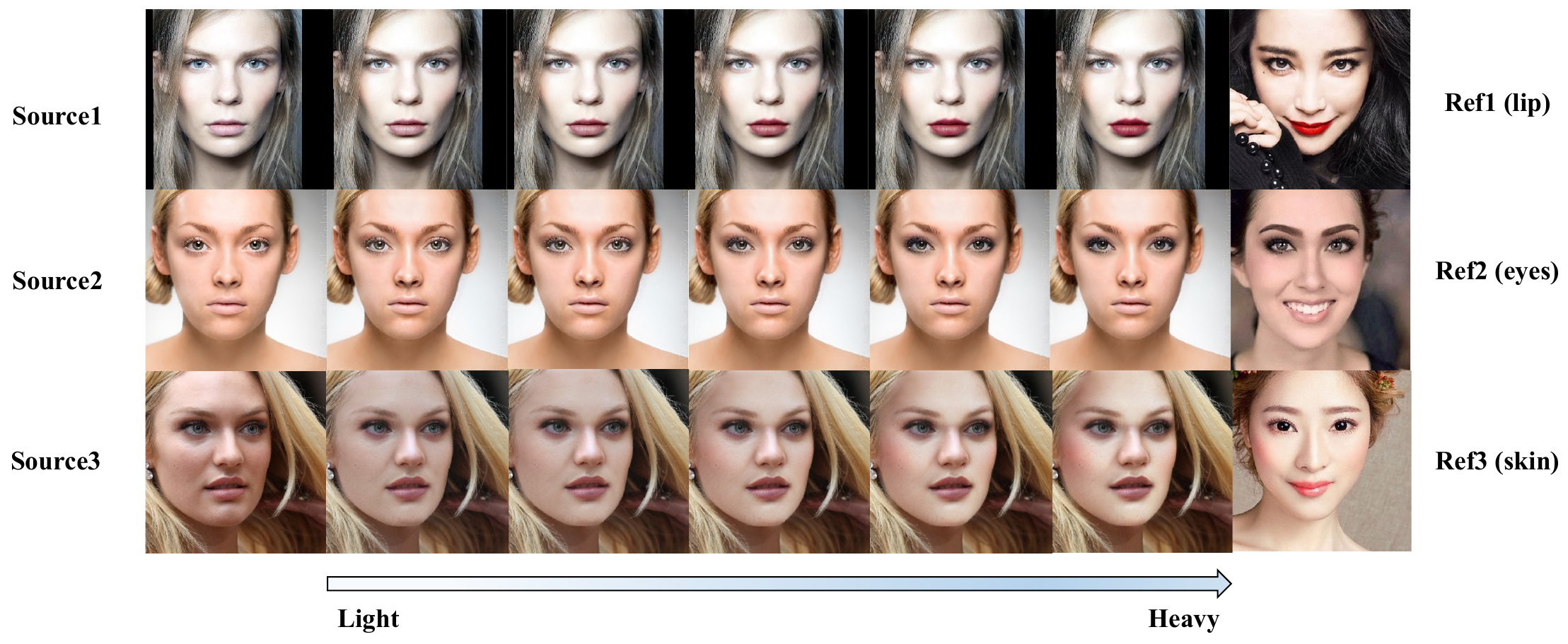}
\caption{\textbf{Partial makeup transfer with interpolation.}} 
\label{8}
\end{figure*}

Moreover, we can simultaneously combine partial makeup transfer and shade-controllable makeup transfer. As demonstrated in Fig. \ref{8}, we independently control the transfer degree for the lips, eyes, and skin. When interpolation is applied to only a specific region, other regions preserve the source identity's features. This is crucial for applying makeup transfer in real-world scenarios, as users often desire to adjust specific parts of the face.

\begin{figure*}
\centering
\setlength{\abovecaptionskip}{0pt}
\includegraphics[width=0.9\textwidth,height=0.16\textheight]{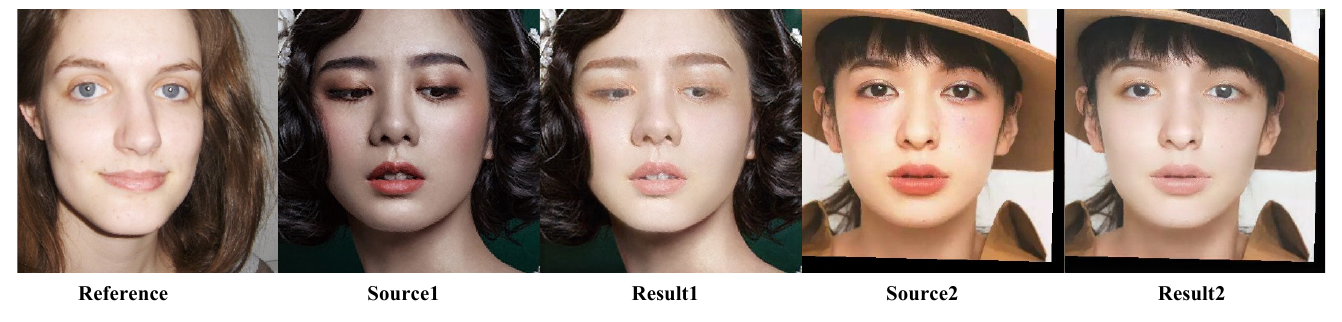}
\caption{\textbf{Makeup removal results.}} 
\label{9}
\end{figure*}

\subsubsection{Makeup Removal} 

By using a non-makeup image as the reference ($y_r$) and a makeup image as the source ($x_s$), we can achieve makeup removal. Fig. \ref{9} displays the makeup removal results, demonstrating that the transfer process is reversible.

\begin{figure*}
\centering
\setlength{\abovecaptionskip}{0pt}
\includegraphics[width=1.0\textwidth,height=0.3\textheight]{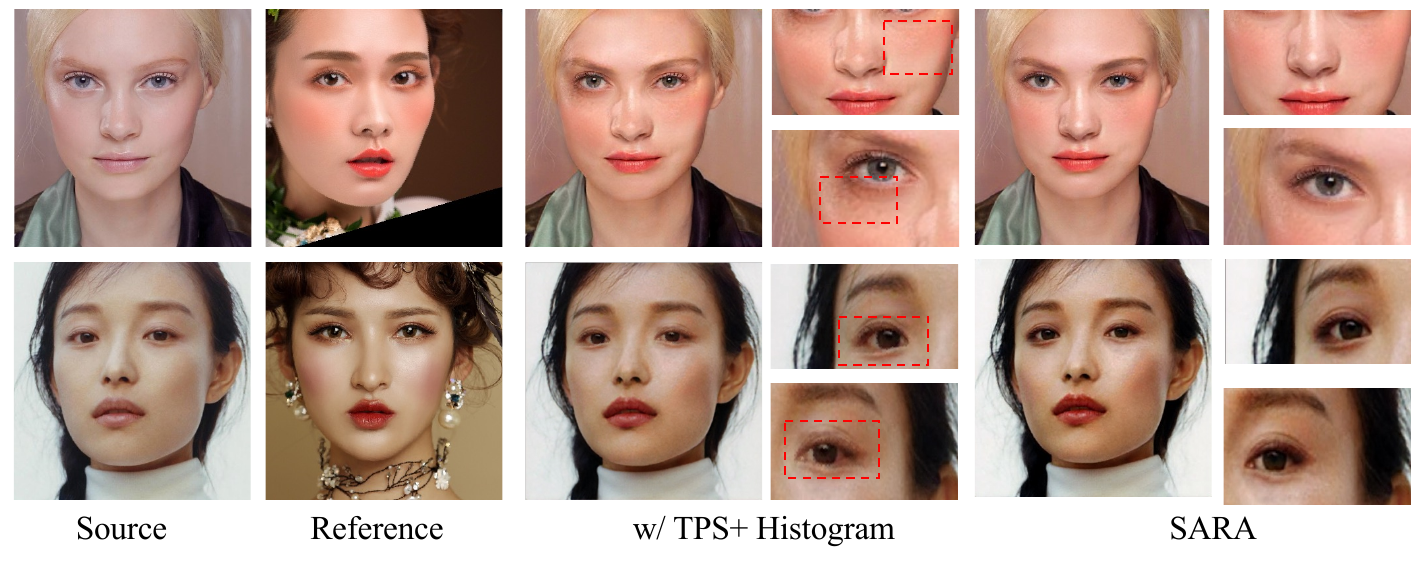}
\caption{\textbf{Ablation study of pseudo ground truth.} $  `w/ TPS+Histogram`$ denotes that the pseudo ground truth is generated using TPS warping and histogram matching, while SARA generates the pseudo ground truth using OT matching and histogram matching. The highlighted red boxes showcase the unnatural results.} 
\label{10}
\end{figure*}

\subsection{Ablation Studies}
\textbf{Effectiveness of pseudo ground truth.} Since there is no ground truth for makeup transfer, we utilize pseudo ground truth for supervision. Histogram matching can constrain the color distributions of the makeup regions. \cite{yang2022elegant} incorporates TPS warping with histogram matching to introduce spatial constraints. However, we argue that TPS with a low degree of freedom is insufficient to warp fine-grained makeup styles under large geometric changes. To address this limitation, we propose using optimal transport matching to establish dense correspondences between distributions. As shown in Fig. \ref{10}, the transfer results in the first row generated by TPS warping exhibit misaligned blush, while SARA can adaptively transfer the blush onto the target face in correct positions. For eye shadow, which requires fine-grained modeling, SARA also achieves better results. In contrast, the eye shadow generated by TPS warping in the second row appears sparse.

\par
\begin{figure*}
\centering
\setlength{\abovecaptionskip}{0pt}
\includegraphics[width=1.0\textwidth,height=0.3\textheight]{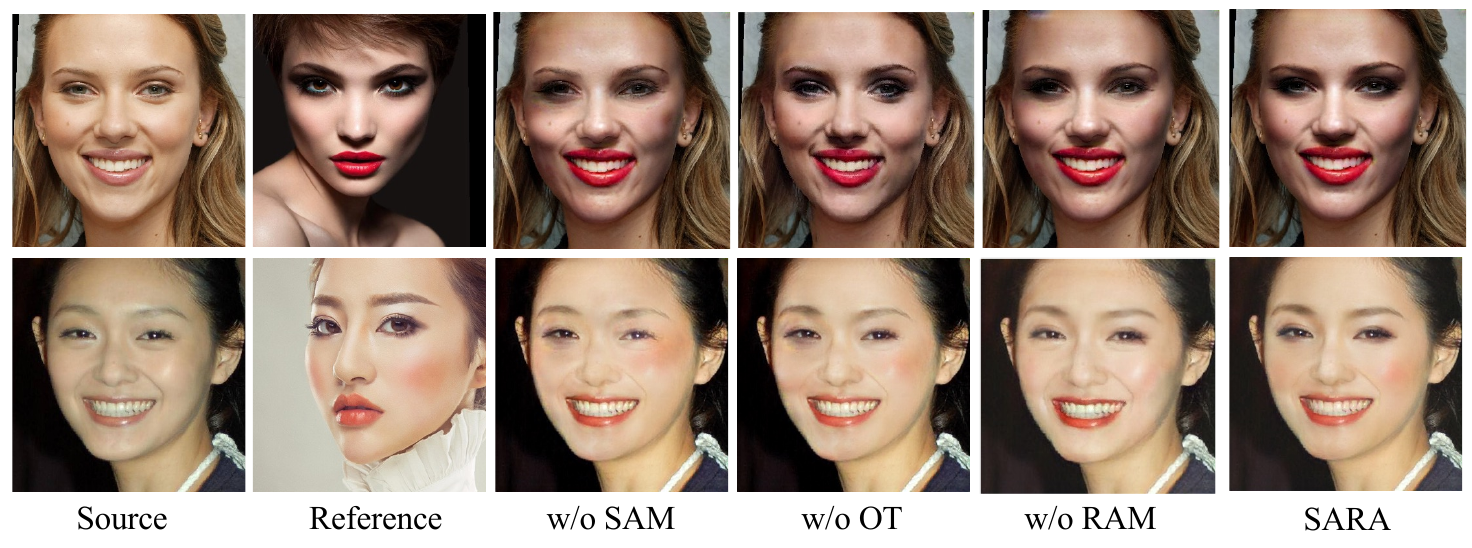}
\caption{\textbf{Ablation study of SAM and RAM.} $`w/o\quad SAM`$ indicates that the transfer is performed only with RAM, while $`w/o\quad OT`$ means that cosine similarity is used for warping in SAM. $`w/o\quad RAM`$ denotes that the transfer is performed only with SAM using optimal transport.} 
\label{11}
\end{figure*}

\textbf{Effectiveness of SAM.} We explicitly construct dense alignment to warp the makeup styles of the reference image under the guidance of semantic information. As shown in the third column of Fig. \ref{11}, when SAM is not used, the transfer is not only inaccurate but also exhibits artifacts. A common approach to establishing dense correspondence is to calculate cosine similarity, but this tends to generate smooth results (Fig. \ref{3}). To address this issue, we propose using unbalanced optimal transport for feature alignment between mismatched semantic regions. The fourth column of Fig. \ref{11} demonstrates the effect of using cosine similarity for dense warping: although it can generally transfer makeup styles, it produces blurry eye shadow and uneven blush.
\par

\textbf{Effectiveness of RAM.} We propose region-adaptive normalization to dynamically combine shape-independent style codes for potential feature loss during the alignment process. As demonstrated in the fifth column of Fig. \ref{11}, the makeup transfer results generated without RAM lack refinement in the eye shadow and blush regions. In contrast, the results produced with RAM exhibit more intricate details and appear more natural.

\par


\section{Conclusion}

We propose SARA, a novel framework for makeup transfer that effectively handles spatial misalignment and enables fine-grained control over the transfer. SARA integrates a semantic-guided alignment module to establish dense correspondence through unbalanced optimal transport, a region-adaptive normalization module to compensate for feature loss via shape-independent style codes, and a makeup fusion module to render detailed results. Extensive experiments show that SARA outperforms existing methods both quantitatively and qualitatively.
\par
\textbf{Acknowledgements}
\par
This work was supported by the National Natural Science Foundation of China (NSFC) 62272172, Guangdong Basic and Applied Basic Research Foundation 2023A1515012920.






\bibliographystyle{IEEEtran}
\bibliography{egbib}

\end{document}